\documentclass[10pt,twocolumn,letterpaper]{article}
\usepackage[square, comma, sort&compress, numbers]{natbib}
\usepackage[pagenumbers]{cvpr} 

\usepackage[utf8]{inputenc} 
\usepackage[T1]{fontenc}    
\usepackage{url}            
\usepackage{booktabs}       
\usepackage{amsfonts}       
\usepackage{nicefrac}       
\usepackage{microtype}      
\usepackage{xcolor}         
\usepackage{tabularx}
\usepackage{subcaption}
\usepackage{multicol}
\usepackage{multirow}
\usepackage{tabularray}
\usepackage{amsmath}
\usepackage{algorithm}%
\usepackage{algorithmicx}%
\usepackage{algpseudocode}
\usepackage{amsfonts}
\usepackage{amssymb}
\usepackage{siunitx}
\usepackage{colortbl}
\usepackage{graphicx}
\usepackage{lipsum}
\usepackage{nicefrac}
\usepackage{tabularray}

\usepackage{pgfplots,gincltex}
\usepgfplotslibrary{groupplots}
\pgfplotsset{compat=1.6}
\usepackage{tikz}
\usetikzlibrary{arrows.meta}

\definecolor{pltblue}{RGB}{174, 199, 232}
\definecolor{pltorange}{RGB}{255, 229, 204}
\definecolor{pltgreen}{RGB}{204, 229, 204}
\definecolor{pltred}{RGB}{229, 204, 204}
\definecolor{pltpurple}{RGB}{239, 218, 230}

\definecolor{tabblue}{HTML}{1f77b4}
\definecolor{taborange}{HTML}{ff7f0e}
\definecolor{tabgreen}{HTML}{2ca02c}
\definecolor{tabred}{HTML}{d62728}
\definecolor{tabpurple}{HTML}{9467bd}

\definecolor{cblue}{RGB}{173, 201, 233}
\definecolor{clblue}{RGB}{222, 234, 246}
\definecolor{corange}{RGB}{255, 152, 67}
\definecolor{lorgange}{RGB}{255, 221, 149}

\definecolor{cvprblue}{rgb}{0.21,0.49,0.74}
\usepackage[pagebackref,breaklinks,colorlinks,allcolors=cvprblue]{hyperref}
\usepackage[capitalize]{cleveref}

\crefname{section}{Sec.}{Secs.}
\Crefname{section}{Section}{Sections}
\Crefname{table}{Table}{Tables}
\crefname{table}{Tab.}{Tabs.}

\def\paperID{0000} 
\def\confName{CVPR}
\def\confYear{2025}

\newcommand{\ourmethod}{RAM }

\title{Restore Anything Model via Efficient Degradation Adaptation}

\author{
\textbf{Bin Ren$^{1,2,3}$\thanks{Work done during visiting at INSAIT Sofia University.}}\quad
\textbf{Eduard Zamfir$^{4}$}\quad
\textbf{Zongwei Wu$^{4}$}\quad
\textbf{Yawei Li$^{5}$}\quad 
\textbf{Yidi Li$^{6}$}\quad \\
\textbf{Danda Pani Paudel$^{3}$}\quad
\textbf{Radu Timofte$^{4}$}\quad
\textbf{Ming-Hsuan Yang$^{7}$}\quad
\textbf{Nicu Sebe$^{2}$}\quad \\
$^1$University of Pisa,
$^2$University of Trento,
$^3$INSAIT Sofia University,
$^4$University of Würzburg, \\
$^5$ETH Z\"urich,
$^6$Taiyuan University of Technology, 
$^7$University of California, Merced
}

\begin{document}
\maketitle

\begin{abstract}
    With the proliferation of mobile devices, the need for an efficient model to restore any degraded image has become increasingly significant and impactful. Traditional approaches typically involve training dedicated models for each specific degradation, resulting in inefficiency and redundancy. More recent solutions either introduce additional modules to learn visual prompts—significantly increasing model size—or incorporate cross-modal transfer from large language models trained on vast datasets, adding complexity to the system architecture.
    In contrast, our approach, termed RAM, takes a unified path that leverages inherent similarities across various degradations to enable both efficient and comprehensive restoration through a joint embedding mechanism—without scaling up the model or relying on large multimodal models. Specifically, we examine the sub-latent space of each input, identifying key components and reweighting them in a gated manner. This intrinsic degradation awareness is further combined with contextualized attention in an X-shaped framework, enhancing local-global interactions. Extensive benchmarking in an all-in-one restoration setting confirms RAM’s SOTA performance, reducing model complexity by approximately 82\% in trainable parameters and 85\% in FLOPs. Our code and models will be publicly available.
\end{abstract}    
\section{Introduction}
\label{sec:introduction}
\begin{figure}[t]
    \centering
    \includegraphics[width=0.97\linewidth]{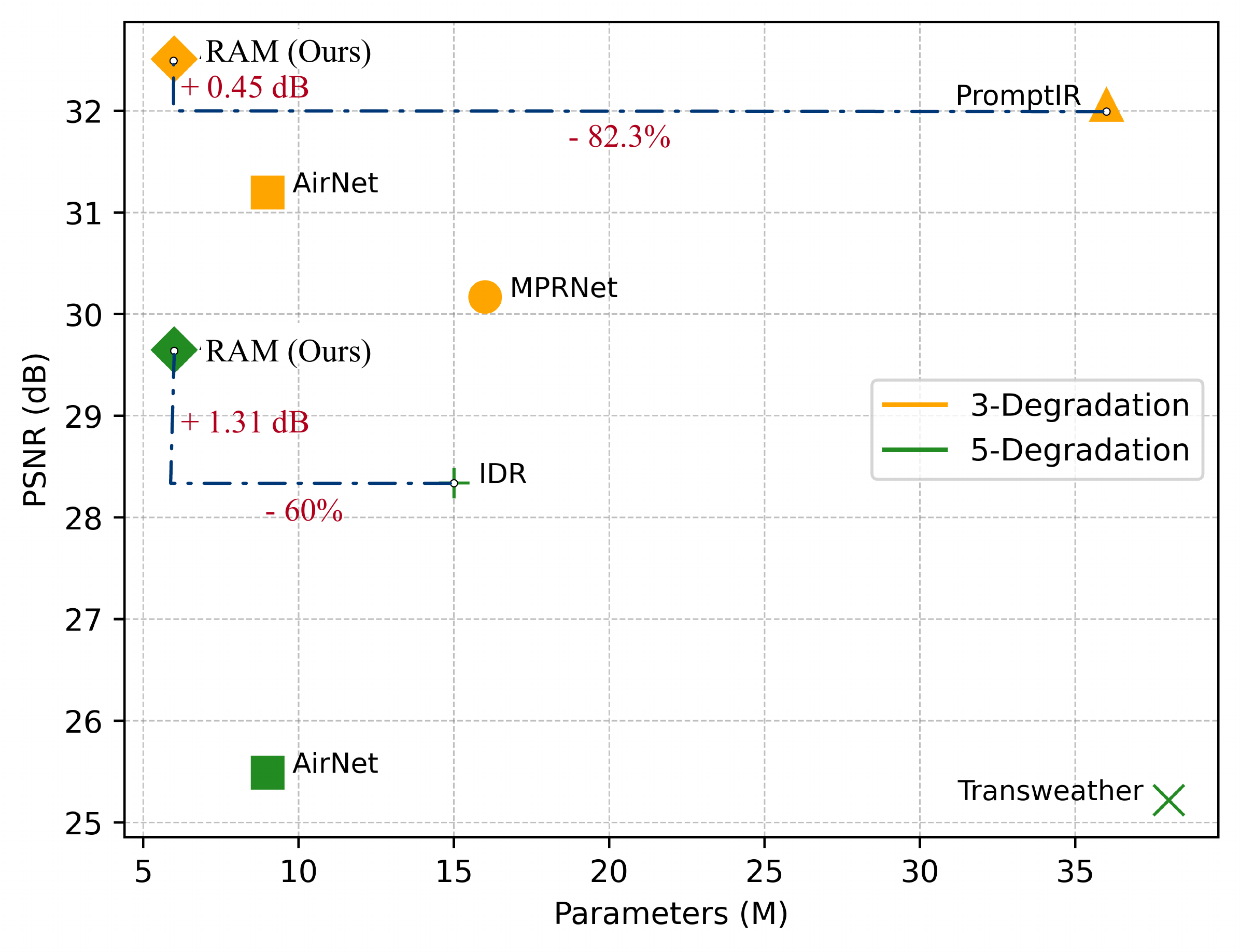}
    \vspace{-2mm}
    \caption{Average PSNR and parameters comparison (\ie, our \ourmethod archives 0.45 dB PSNR improvement with around 82\% parameters reduced compared to the SOTA PromptIR~\cite{potlapalli2023promptir})}
    \label{fig:teaser}
    \vspace{-1em}
\end{figure}

Image restoration (IR) is a fundamental task within low-level computer vision, aiming to enhance the quality of images affected by numerous factors, including noise~\cite{zhang2017beyond,zhang2017learning,zhang2018ffdnet}, blur~\cite{shan2008high,pan2014deblurring,tsai2022banet,kong2023efficient}, compression artifacts~\cite{dong2015compression,jiang2021towards,ehrlich2020quantization}, adverse weather conditions~\cite{cai2016dehazenet,berman2016non,li2018single,li2016rain,yang2017deep,li2019heavy}, and other forms of distortion. It supports downstream vision tasks such as object detection, recognition, and tracking~\cite{sezan1982image,molina2001image,ren2024ninth,cai2023spsd,liu2020grouped}.
Despite significant advances in recent years, existing IR methods struggle to efficiently handle complex distortions while preserving or recovering essential image details~\cite{li2023efficient,ren2024ninth}. The state-of-the-art methods often impose a substantial computational burden, limiting their applicability, especially on edge and mobile devices. Thus, there is a strong demand for an efficient model to handle any type of image degradation.

The complexity of existing restoration models~\cite{liang2021swinir,zamir2022restormer,zamir2021multi,wang2022uformer,chen2022simple} arises mainly from the diversity of the types of degradation. Consequently, many existing methods are tailored for a specific type of degradation and application with limited generalization to others. A versatile system would require the integration of multiple specialized models, resulting in a complex and heavy framework. Some recent work has shown that a single architecture can address multiple types of degradation, but these approaches lack parameter unification, leading to multiple checkpoints, each for a specific task\cite{zamir2022restormer,chen2022simple,wang2022uformer,chen2022cross}. Although this reduces system complexity, it compromises efficiency due to the increased number of checkpoint sets.

Recent studies have focused on achieving both architectural and parameter unification~\cite{li2022airnet,potlapalli2023promptir,zamfir2024efficient,zhang2023ingredient,liu2022tape}. Diffusion-based methods have demonstrated improved image generation capabilities~\cite{ren2023multiscale,jiang2023autodir,zhao2024denoising}. To improve the controllability of the network, other approaches treat each modality as an independent visual prompt, which guides the network toward the target application~\cite{wang2023promptrestorer,potlapalli2023promptir,li2023prompt}. Another group of works uses text as an intermediate representation to unify all degradations, using textual instructions to prompt the network~\cite{luo2023controlling,conde2024instruct}. Despite promising results, these networks still require significant computational resources, with a huge number of parameters and low inference speed.

In this paper, we propose a novel perspective that treats image restoration as a unified problem, arguing that \textit{despite degradation-specific characteristics, all restoration tasks share underlying similarities}.
Since all types of restoration require an understanding of the global context to preserve structural integrity, while also being sensitive to local details and capable of reducing noise affecting quality. 
Learning these general techniques is not limited to one specific task, but is generalizable and applicable across various types of degradation. 
This aligns with the motivation behind recent advancements in large language models (LLMs), which have demonstrated the potential to learn a joint embedding or utilize a single network for multiple tasks. 
Unlike LLMs that achieve unified parameters through learning from large data priors with substantial models, we focus on deeply extracting the inherent details of a single image to build correlations across degradations. 
Our approach enables us to achieve comparable performance without the need for specific modules and unique parameter sets, resulting in significantly fewer parameters and enabling emergent alignment in image restoration.

To efficiently achieve such a unification, we propose a novel lightweight attention mechanism that optimally intertwines global context and local details. Specifically, we introduce a gated local block that learns degradation-aware details, aiming to discern different degradation types and handle nuanced differences at a detailed level. Technically, for each feature, we analyze its detailed components, comprising the ego, shifted, and scaled parts. Leveraging the network's self-adaptation capability, we reweight and gate the contribution of each key component intelligently and adaptively, reshaping the output into a locally optimized form with enhanced awareness of its degradation type. This local awareness is then integrated with transformer attention in an X-shaped scheme to maximize the local-global intertwining. Note that for feature modeling, we split the input feature into subparts and process them in a sub-latent space before final aggregation. This strategy not only enables inner-feature interaction but also reduces computational cost, making our model both effective and efficient.

The contributions of this work are:
\begin{itemize}
    \item We propose \ourmethod, a single model capable of efficiently handling any type of degradation. Compared to the SOTA all-in-one counterparts, our model reduces the computational cost by $ 85.6\%$ while delivering superior overall performance.
    \item We introduce a novel local-global gated intertwining within the feature modeling block. This design deeply explores the intrinsic characteristics of each degradation type as guidance, without the need for a degradation-specific design, leading to a cohesive embedding for comprehensive IR.
    \item Extensive comparisons across different IR tasks validate the effectiveness and efficiency of our proposed method. We hope our efficient network can serve as a baseline and inspire future research in the domain.
\end{itemize}
\section{Related Work}
\label{sec:related-work}
\noindent{\textbf{Image Restoration (IR)}.}
Image restoration aims to solve a highly ill-posed problem by reconstructing high-quality images from their degraded counterparts. Due to its importance, IR has been applied to various applications~\cite{richardson1972bayesian,banham1997digital,li2023lsdir,zamfir2024details}. Initially, IR was addressed through model-based solutions involving the search for solutions to specific formulations. However, learning-based approaches have gained much attention with the significant advancements in deep neural networks. Numerous approaches have been developed, including regression-based~\cite{lim2017enhanced,lai2017deep,liang2021swinir,chen2021learning,li2023efficient,zhang2024transcending} and generative model-based pipelines~\cite{gao2023implicit,wang2022zero,luo2023image,yue2023resshift,zhao2024denoising,liu2023spatio} that built on convolutional~\cite{dong2015compression,zhang2017learning,zhang2017beyond,wang2018recovering}, MLP~\cite{tu2022maxim}, state space mode~\cite{guo2024mambair,zhu2024vision,mamba,mamba2}, or vision transformers-based (ViTs) architectures~\cite{liang2021swinir,li2023efficient,zamir2022restormer,ren2023masked,dosovitskiy2020image}. 
Although state-of-the-art methods have achieved promising performance, the mainstream IR solutions still focus on addressing single degradation tasks such as denoising~\cite{zhang2017learning,zhang2019residual}, dehazing~\cite{ren2020single,wu2021contrastive}, deraining~\cite{jiang2020multi,ren2019progressive}, deblurring~\cite{kong2023efficient,ren2023multiscale}, and so on. In this work, the proposed method is built based on the regression-based pipeline with ViT architecture. 

\begin{figure*}[!t]
    \centering
    \includegraphics[width=1.0\linewidth]{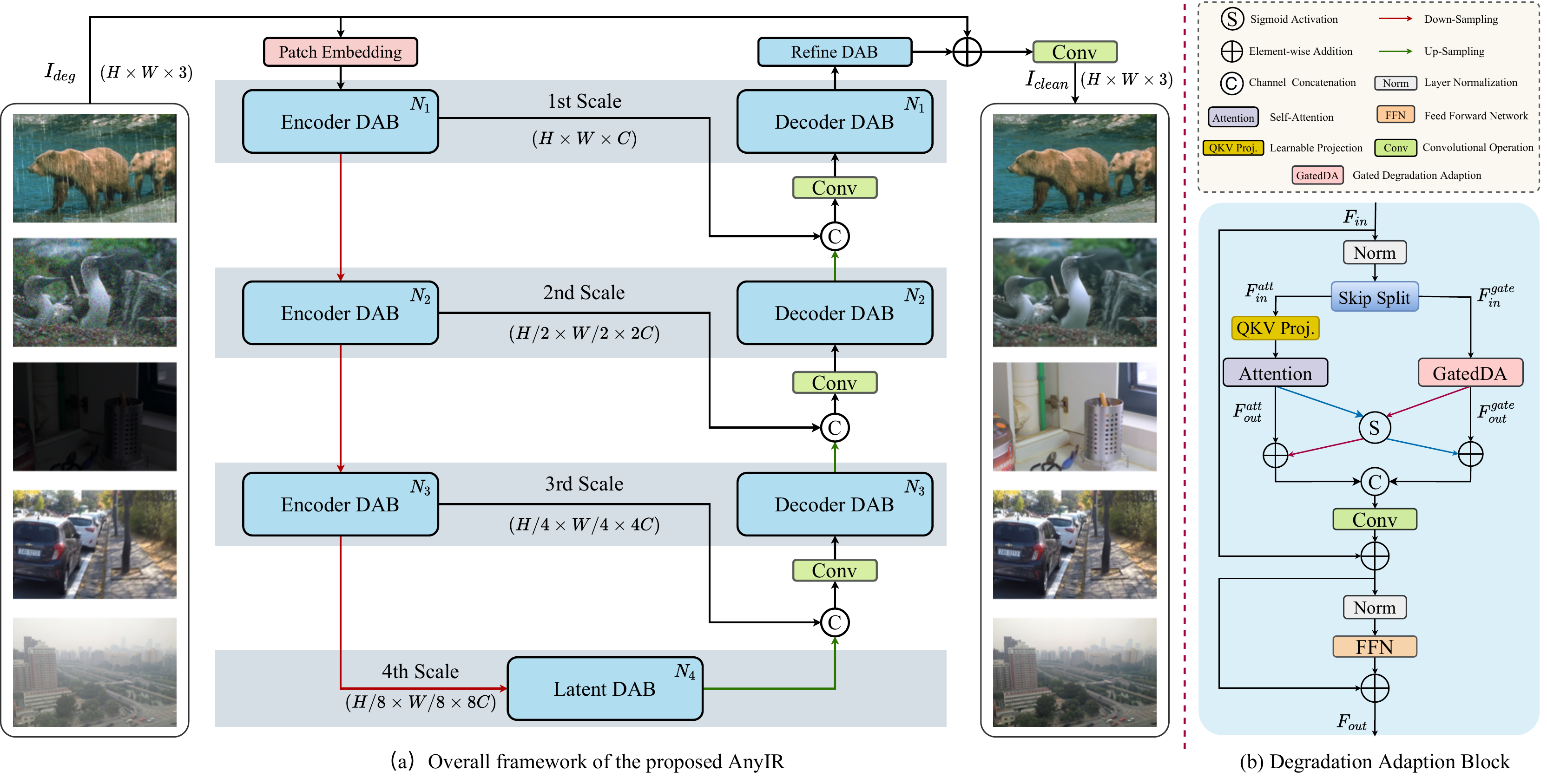}
        \vspace{-6mm}
    \caption{(a) Overall framework of the proposed \ourmethod: a U-shaped structure which mainly consists of a convolutional patch embedding, a U-shape encoder-decoder main body, and an extra refined block. (b) Structure of each degradation adaptation block (DAB)}
    \label{fig:framework}
    \vspace{-1em}
\end{figure*}

\paragraph{One for Any Image Restoration.}
Training a task-specific model to tackle a single type of degradation is effective, but practical implementation is challenging due to the need for separate models for each degradation type. In reality, images often suffer from multiple degradations and artifacts, making it difficult to address them individually. Task-specific solutions require significant computing and storage resources, substantially increasing their environmental impact. To address this, the emerging All-in-One image restoration field uses a single-blind restoration model to handle multiple degradation types simultaneously.
Specifically, AirNet~\cite{li2022all} accomplishes blind All-in-One image restoration by leveraging contrastive learning to derive degradation representations from corrupted images, which are then used to reconstruct the clean image. Following this, IDR~\cite{zhang2023ingredient} addresses All-in-One image restoration by decomposing image degradations into their fundamental physical principles and employing a two-stage meta-learning approach. 
More recently, the prompt-based paradigm~\cite{potlapalli2023promptir,wang2023promptrestorer,li2023prompt} has been introduced, incorporating an additional visual prompt learning module. This approach has been validated as effective in guiding a single trained model to better restore images with multiple types of degradation by leveraging the learned discriminative capabilities of the visual prompts. 
Built upon the prompt-based paradigm, there also some recent works propose modeling the visual prompt either from the frequency perspective~\cite{cui2024adair} or by posting more complex architecture designs with extra dataset~\cite{dudhane2024dynamic,zamfir2024efficient}. 
However, using visual prompt modules often leads to longer training times and reduced efficiency. Instead, our work enhances the model's ability to capture representative information without the burden of heavy and complex prompt modules.
\section{Proposed \ourmethod}
\label{sec:method}
\paragraph{Overview.} 
In this section, we present our efficient All-in-One restoration method, termed \ourmethod. The overall framework is illustrated in \cref{fig:framework}. At a macro level, \ourmethod employs a U-shaped network architecture~\cite{ronneberger2015unet} with four levels. Each level incorporates $N_{i}, i\in [1,2,3,4]$ instances of our novel Degradation Adaptation Blocks (DAB)~\ref{subsec:adab}, and each DAB is built upon the proposed gated degradation adaption (GatedDA) module (Sec.~\ref{subsec:gatedDA}).
Initially, a convolutional layer first extracts shallow features from the degraded input, creating a patch embedding of \( H \times W \times C \). Subsequently, each U-Net level doubles the embedding dimension and halves the spatial resolution. Skip connections transfer information from the encoder stage to the decoder. At the decoder, we merge these features with the signal from the previous decoding stage using a linear projection. Lastly, a global residual connection links the input image to the output, preserving high-frequency details and producing the restored image.

\subsection{Degradation Adaptation Block}
\label{subsec:adab}
Our objective is to minimize the overall computational complexity while proficiently encapsulating the spatial data crucial for image restoration. 
Neural meta-architectures have gained significant attention in computer vision~\cite{yu2022metaformer,liu2022convnet,yu2024inceptionnext}, with convolutional-based networks achieving performance comparable to Transformers. 
Recently, Mamba-based models~\cite{guo2024mambair,mamba,mamba2} have revisited the concept of gated blocks, simplifying the meta-architecture into an integration of a spatial mixer convolution and a Feed-forward network. 
For achieving an efficient image restorer, we propose a hybrid architecture, combining a parameter-reduced Attention layer for global feature learning, while a gated convolutional block aggregates degradation-specific local details.

The detailed structure of the proposed DAB is illustrated in Fig.~\ref{fig:framework} (b). Given an input feature tensor \( F_{in} \in \mathbb{R}^{H \times W \times C} \), where \( H \), \( W \), and \( C \) denote the height, width, and channel dimensions, respectively, we employ a selective channel-wise partitioning strategy defined as follows:
\begin{equation}
\begin{aligned}
    F_{in}^{att} &= \{ F_{in}^{(2k-1)} \mid k \in \mathbb{Z}^+, k \leq \frac{C}{2} \}, \\
    F_{in}^{gate} &= \{ F_{in}^{(2k)} \mid k \in \mathbb{Z}^+, k \leq \frac{C}{2} \},
\end{aligned}
\label{eqn:1}
\end{equation}
where \( F_{in}^{att} \) and \( F_{in}^{gate} \in \mathbb{R}^{H \times W \times \nicefrac{C}{2}} \) denote the sub-features derived from interleaved sampling along the channel dimension. This setup provides two distinct pathways: \( F_{in}^{att} \), focusing on attention modulation, and \( F_{in}^{gate} \), specializing in gated transformation. 
This structured skip-split offers two notable benefits, \ie, 
\textit{(i)} Significant Complexity Reduction: By reducing the effective channel dimension in each pathway, the overall model complexity and parameter count are notably minimized, which is particularly beneficial for the self-attention layers in Vision Transformers (ViTs)~\cite{dosovitskiy2020image}.
\textit{(ii)} Improved Information Retention: Unlike conventional half-split methods, the skip-split preserves feature diversity by uniformly sampling the original channels, ensuring each subset retains comprehensive information from \( F_{in} \) without loss of essential channel-wise details. 

\begin{figure}[!t]
    \centering
    \includegraphics[width=1.0\linewidth]{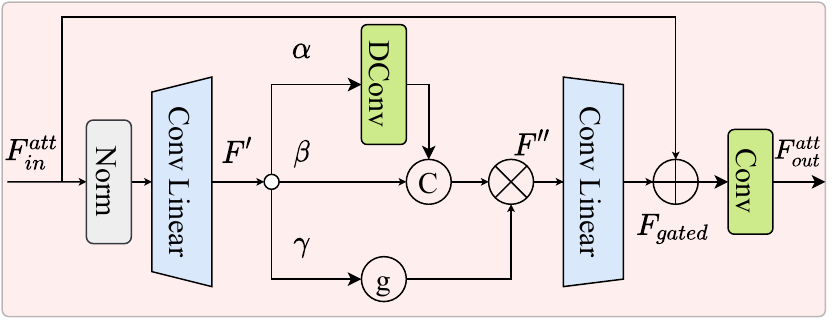}
    \caption{Structure of our GatedDA. $\oplus$, $\textcircled{c}$, $\textcircled{g}$, and $\otimes$ denote the element-wise addition, channel-wise concatenation, GELU~\cite{hendrycks2016gaussian} activation, and element-wise multiplication, respectively.}
    \label{fig:GatedDA}
\end{figure}

To capture the complex dependencies inherent to global degradation, we employ a multi-depth convolution head attention mechanism~\cite{zamir2022restormer,potlapalli2023promptir} on the feature subset \( F_{in}^{att} \), resulting in \( F_{out}^{att} \). This attention approach is particularly effective for image restoration, where degradation is often non-uniform and shaped by various intricate factors.
Specifically, \( F_{in}^{att} \) is transformed into Query (\( Q \)), Key (\( K \)), and Value (\( V \)) matrices, defined as:
$
Q = F_{in}^{att} \mathbf{W}_{qry}, \quad K = F_{in}^{att} \mathbf{W}_{key}, \quad V = F_{in}^{att} \mathbf{W}_{val},
$
where \( \mathbf{W}_{qry} \), \( \mathbf{W}_{key} \), and \( \mathbf{W}_{val} \) are learnable weights. The output \( F_{out}^{att} \) is then calculated as:
\begin{equation}
    F_{out}^{att} = \sum_i \texttt{Softmax} \left( \frac{QK^{\top}}{\sqrt{d}} \right)_i V_{i,j},
\label{eqn:attention_output}
\end{equation}
where \(\sqrt{d}\) is a scaling factor to normalize attention scores, stabilizing gradients and aiding convergence. This attention mechanism excels at modeling long-range dependencies across the feature space, a vital capability for image restoration where degradation—such as noise, blur, and artifacts—exhibits spatial correlations yet varies across the image~\cite{liang2021swinir,zamir2022restormer,li2023efficient,chen2023activating}. By leveraging this global modeling, the framework dynamically identifies and emphasizes degraded regions, preserving fine details essential for high-quality restoration.

Meanwhile, the proposed GatedDA operation (see \cref{subsec:gatedDA}) is applied to \( F_{in}^{gate} \), producing \( F_{out}^{gate} \). This approach combines the strengths of attention and convolution: the attention module captures global structural cues essential for overall image context, while the convolutional pathway captures local dependencies crucial for refining details in low-level vision tasks~\cite{liang2021swinir,chen2023dat,chen2023rgt}. To enable effective interaction between these global and local features, we introduce a cross-feature update:
\begin{equation}
\begin{aligned}
    F_{out}^{ag} &= F_{out}^{att} + \sigma \left( F_{out}^{gate} \right), \\
    F_{out}^{ga} &= F_{out}^{gate} + \sigma \left( F_{out}^{att} \right),
\end{aligned}
\label{eqn:cross_update}
\end{equation}
where \( \sigma(\cdot) \) is the Sigmoid activation, enhancing each feature map with complementary information. Through this update, \( F_{out}^{att} \) gains local detail sensitivity, while \( F_{out}^{gate} \) benefits from structural context. The fused representation \( F_{fuse} \) is then computed by concatenating \( F_{out}^{att} \) and \( F_{out}^{gate} \), followed by a convolution with learnable weights \( \mathbf{W}_{fuse} \) and a residual connection to the input:
\begin{equation}
    F_{fuse} = \operatorname{Conv}_{\mathbf{W}_{fuse}}(\operatorname{concat}(F_{out}^{att}, F_{out}^{gate})) + F_{in}.
\end{equation}
This fusion step integrates global and local information, enhancing feature richness for the final restoration. Next, layer normalization, a feed-forward network (FFN), and a residual connection are applied, yielding \( F_{out} \):
\begin{equation}
    F_{out} = \operatorname{FFN}(\operatorname{Norm}(F_{fuse})) + F_{fuse}.
\end{equation}
This sequence stabilizes feature distributions and boosts expressive capacity, effectively combining global and local cues for high-fidelity, detailed image restoration.

\subsection{Gated Degradation Adaption}
\label{subsec:gatedDA}

\begin{algorithm}[t]
\caption{Gated Degradation Adaption}
\label{alg:gated_degradation_adaption}
\begin{algorithmic}[1]
    \Require Input $F_{in}^{\text{gate}} \in \mathbb{R}^{C \times H \times W}$, initial temperature $\tau$
    \Ensure Output $F_{out}^{\text{gate}} \in \mathbb{R}^{C \times H \times W}$
    
    \State $\hat{F} \gets \text{Norm}(F_{in}^{\text{gate}})$ \hfill // Normalize
    \State $\mu, \sigma \gets \frac{1}{HW} \sum \hat{F}, \sqrt{\frac{1}{HW} \sum (\hat{F} - \mu)^2}$ \hfill // Mean, std
    \State $\alpha \gets \sigma(\mu + \sigma)$ \hfill // Temp adjust
    \State $\tau_{\text{adj}} \gets \tau \cdot \alpha$
    
    \State $F' \gets \hat{F} \mathbf{W}_{\text{exp}}$ \hfill // Channel expand
    \State $\gamma, \beta, \alpha \gets \text{split}(F')$ \hfill // Split $\gamma$, $\beta$, $\alpha$
    \State $\alpha' \gets (\alpha \mathbf{W}_{\text{depth}}) \cdot \tau_{\text{adj}}$ \hfill // Depthwise conv
    
    \State $F_{\text{gated}} \gets \sigma(\gamma) \cdot \text{concat}(\beta, \alpha') \mathbf{W}_{\text{gate}}$ \hfill // Gate combine
    \State $F_{out}^{\text{gate}} \gets (F_{\text{gated}} + F_{in}^{\text{gate}}) \mathbf{W}_{\text{proj}}$ \hfill // Residual, proj
    
    \State \Return $F_{out}^{\text{gate}}$
\end{algorithmic}
\end{algorithm}

\begin{table*}[t]
    \centering
    \footnotesize
    \fboxsep0.75pt
    \setlength{\extrarowheight}{0.8pt}
    \caption{\textit{Comparison to state-of-the-art on three degradations.} PSNR (dB, $\uparrow$) and SSIM ($\uparrow$) metrics are reported on the full RGB images. \textcolor{tabred}{\textbf{Best}} and \textcolor{tabblue}{\textbf{second best}} performances are highlighted. Our method sets a new state-of-the-art on average across all benchmarks while being significantly more efficient than prior work. ‘-’ represents unreported results.}
    \vspace{-3mm}
    \label{tab:exp:3deg}
    \begin{tabularx}{\textwidth}{X*{15}{c}}
    \toprule
     \multirow{2}{*}{Method} & \multirow{2}{*}{Params.} 
     & \multicolumn{2}{c}{\textit{Dehazing}} & \multicolumn{2}{c}{\textit{Deraining}} & \multicolumn{6}{c}{\textit{Denoising}} 
     & \multicolumn{2}{c}{\multirow{2}{*}{Average}}  \\
     \cmidrule(lr){3-4} \cmidrule(lr){5-6} \cmidrule(lr){7-12} 
     && \multicolumn{2}{c}{SOTS} & \multicolumn{2}{c}{Rain100L} & \multicolumn{2}{c}{BSD68\textsubscript{$\sigma$=15}} & \multicolumn{2}{c}{BSD68\textsubscript{$\sigma$=25}} & \multicolumn{2}{c}{BSD68\textsubscript{$\sigma$=50}} &  \\
     \midrule
        \rowcolor{gray!10} BRDNet~\cite{tian2000brdnet} & - & 23.23 & {.895} & 27.42 & {.895} & 32.26 & {.898} & 29.76 & {.836} & 26.34 & {.693} & 27.80 & {.843} \\
        LPNet~\cite{gao2019dynamic} & - & 20.84 & {.828} & 24.88 & {.784} & 26.47 & {.778} & 24.77 & {.748} & 21.26 & {.552} & 23.64 & {.738} \\
        \rowcolor{gray!10} FDGAN~\cite{dong2020fdgan}  & - & 24.71 & {.929} & 29.89 & {.933} & 30.25 & {.910} & 28.81 & {.868} & 26.43 & {.776} & 28.02 & {.883} \\
        DL~\cite{fan2019dl} & 2M & 26.92 & {.931} & 32.62 & {.931} & 33.05 & {.914} & 30.41 & {.861} & 26.90 & {.740}  & 29.98 & {.876}\\
        \rowcolor{gray!10} MPRNet~\cite{zamir2021pmrnet}  & 16M & 25.28 & {.955} & 33.57 & {.954} & 33.54 & {.927} & 30.89 & {.880} & 27.56 & {.779} & 30.17 & {.899} \\
        AirNet~\cite{li2022airnet} & 9M & 27.94 & {.962} & 34.90 & {.967} & 33.92 & \textcolor{tabblue}{\textbf{.933}} & 31.26 & {.888} & 28.00 & {.797} & 31.20 & {.910} \\
        \rowcolor{gray!10} PromptIR~\cite{potlapalli2023promptir} & 36M & \textcolor{tabblue}{\textbf{30.58}} & \textcolor{tabblue}{\textbf{.974}} & \textcolor{tabblue}{\textbf{36.37}} &  \textcolor{tabblue}{\textbf{.972}} & \textcolor{tabred}{\textbf{33.98}} &  \textcolor{tabred}{\textbf{.933}} & \textcolor{tabred}{\textbf{31.31}} & \textcolor{tabblue}{\textbf{.888}} & \textcolor{tabred}{\textbf{28.06}} &  \textcolor{tabred}{\textbf{.799}} & \textcolor{tabblue}{\textbf{32.06}} &  \textcolor{tabblue}{\textbf{.913}} \\
        \midrule
        \ourmethod (\textit{Ours}) & \textcolor{tabred}{\textbf{6M}} & \textcolor{tabred}{\textbf{31.38}} &  \textcolor{tabred}{\textbf{.979}} & \textcolor{tabred}{\textbf{37.90}} &  \textcolor{tabred}{\textbf{.981}} & \textcolor{tabblue}{\textbf{33.95}} &  \textcolor{tabred}{\textbf{.933}} & \textcolor{tabblue}{\textbf{31.30}} & \textcolor{tabred}{\textbf{.889}} & \textcolor{tabblue}{\textbf{28.04}} & \textcolor{tabblue}{\textbf{.797}} & \textcolor{tabred}{\textbf{32.51}} &  \textcolor{tabred}{\textbf{.916}} \\
        \midrule
        \multicolumn{14}{c}{Methods with the assistance of vision language, multi-task learning, natural language prompts, and multi-modal control} \\
        \midrule 
        \rowcolor{gray!10} DA-CLIP~\cite{luo2023controlling} & 125M & 29.46 & .963 & 36.28 & .968 &  30.02 & .821 & 24.86 & .585 & 22.29 & .476 & - & - \\
        Art$_{PromptIR}$~\cite{wu2024harmony} & 36M & 30.83 &.979 & 37.94 & .982 &34.06 & .934 & 31.42 & .891 & 28.14 & .801 & 32.49 & .917 \\
        \rowcolor{gray!10} InstructIR-3D~\cite{conde2024high} & 16M & 30.22 &.959 & 37.98 & .978 & 34.15& .933 &31.52&.890& 28.30&.804 &32.43&.913 \\
        UniProcessor~\cite{duan2025uniprocessor} & - & 31.66 & .979 &  38.17 & .982 & 34.08 &.935 &  31.42 & .891 & 28.17 & .803 & 32.70 & .918  \\
     \bottomrule
    \end{tabularx}
    \vspace{-4mm}
\end{table*}

To capture local degradation-aware details, we leverage the selective properties of gated convolution. Given an input feature map \( F_{in}^{\text{gate}} \in \mathbb{R}^{C \times H \times W} \), a layer normalization is first applied to stabilize feature distributions, followed by a 1\(\times\)1 convolution that expands the channel dimension to \( \text{hidden} = r_{\text{expan}} \cdot C \), where \( r_{\text{expan}} \) is the expansion ratio, as shown in Alg.~\ref{alg:gated_degradation_adaption}. 
To adaptively respond to varying intensities, we introduce a temperature adjustment mechanism. Based on the input’s mean and standard deviation, the initial temperature \( \tau \) is modulated as \( \tau_{\text{adj}} = \tau \cdot \alpha \), where \( \alpha \) serves as a dynamic scaling factor (see steps 2–4 of Algorithm~\ref{alg:gated_degradation_adaption}). This temperature adjustment enhances the model’s sensitivity to subtle degradation patterns, promoting detailed feature capture.
The expanded feature \( F' \) is then split into three parts: \(\gamma\), \(\beta\), and \(\alpha\) (as shown in step 6). Here, \(\alpha\) undergoes depthwise convolution with \( \tau_{\text{adj}} \) to capture spatial details, \(\beta\) retains the original information, and \(\gamma\) is activated with GELU~\cite{hendrycks2016gaussian} to enable a non-linear gated selection mechanism. These components are recombined and modulated, selectively emphasizing critical features (step 7).
Finally, \( F'' \) is projected back to the original channel dimension and combined with \( F_{in}^{\text{gate}} \) through a skip connection to prevent information loss. A final 1\(\times\)1 convolution fuses the degradation-adapted features with the input, yielding the output \( F_{out}^{\text{gate}} \) (steps 8–9 in Algorithm~\ref{alg:gated_degradation_adaption}).
In summary, the GatedDA block dynamically adjusts temperature and gating mechanisms to adaptively capture degradation-sensitive features, balancing global structure with fine detail. Further analysis is available in the \textit{Supplementary Materials} (\ie, \textit{Supp. Mat.}).
\section{Experiments and Comparisons}
\label{sec:experiments}
We conduct experiments adhering to the protocols of prior general image restoration works \cite{potlapalli2023promptir,zhang2023ingredient} under two settings: 
\textit{(i) All-in-One} and 
\textit{(ii) Single-task}. 
In the All-in-One setting, a unified model is trained to handle multiple degradation types, considering \textit{three} and \textit{five} distinct degradations. In the Single-task setting, separate models are trained for each specific restoration task.
The implementation details and the dataset introduction are provided in our \textit{Supp. Mat.}

\subsection{State of the Art Comparison}

\begin{table*}[t]
    \centering
    \footnotesize
    \fboxsep0.75pt
    \setlength{\extrarowheight}{0.2pt}
    \caption{\textit{Comparison to state-of-the-art on five degradations.} PSNR (dB, $\uparrow$) and SSIM ($\uparrow$) metrics are reported on the full RGB images with $(^\ast)$ denoting general image restorers, others are specialized all-in-one approaches. \textcolor{tabred}{\textbf{Best}} and \textcolor{tabblue}{\textbf{second best}} performances are highlighted.}
    \vspace{-3mm}
    \label{tab:exp:5deg}
    \begin{tabularx}{\textwidth}{X*{15}{c}}
    \toprule
     \multirow{2}{*}{Method} & \multirow{2}{*}{Params.} 
     & \multicolumn{2}{c}{\textit{Dehazing}} & \multicolumn{2}{c}{\textit{Deraining}} & \multicolumn{2}{c}{\textit{Denoising}} 
     & \multicolumn{2}{c}{\textit{Deblurring}} & \multicolumn{2}{c}{\textit{Low-Light}} & \multicolumn{2}{c}{\multirow{2}{*}{Average}}  \\
     \cmidrule(lr){3-4} \cmidrule(lr){5-6} \cmidrule(lr){7-8} \cmidrule(lr){9-10} \cmidrule(lr){11-12}
     && \multicolumn{2}{c}{SOTS} & \multicolumn{2}{c}{Rain100L} & \multicolumn{2}{c}{BSD68\textsubscript{$\sigma$=25}} 
     & \multicolumn{2}{c}{GoPro} & \multicolumn{2}{c}{LOLv1} &  \\
     \midrule
     
    \rowcolor{gray!10} NAFNet$^\ast$~\cite{chen2022simple} & 17M & 25.23 & {.939} & 35.56 & {.967} & 31.02 & {.883} & 26.53 & {.808} & 20.49 & {.809} & 27.76 & {.881} \\
    DGUNet$^\ast$~\cite{mou2022deep} & 17M & 24.78 & {.940} & 36.62 & {.971} & 31.10 & {.883} & 27.25 & {.837} & 21.87 & {.823} & 28.32 & {.891} \\
    \rowcolor{gray!10} SwinIR$^\ast$~\cite{liang2021swinir} & 1M & 21.50 & {.891} & 30.78 & {.923} & 30.59 & {.868} & 24.52 & {.773} & 17.81 & {.723} & 25.04 & {.835} \\ 
    Restormer$^\ast$~\cite{zamir2022restormer} & 26M & 24.09 & {.927} & 34.81 & {.962} & 31.49 & {.884} & 27.22 & {.829} & 20.41 & {.806} & 27.60 & {.881} \\
    \rowcolor{gray!10} MambaIR$^\ast$~\cite{guo2024mambair} & 27M & 25.81 & .944 & 36.55 & .971 & 31.41 & .884 & 28.61 & .875 & 22.49 & .832 & 28.97 &.901 \\
    \midrule
    DL~\cite{fan2019dl} & 2M & 20.54 & {.826} & 21.96 & {.762} & 23.09 & {.745} & 19.86 & {.672} & 19.83 & {.712} & 21.05 & {.743} \\
    \rowcolor{gray!10} Transweather~\cite{valanarasu2022transweather} & 38M & 21.32 & {.885} & 29.43 & {.905} & 29.00 & {.841} & 25.12 & {.757} & 21.21 & {.792} & 25.22 & {.836} \\
    TAPE~\cite{liu2022tape} & 1M & 22.16 & {.861} & 29.67 & {.904} & 30.18 & {.855} & 24.47 & {.763} & 18.97 & {.621} & 25.09 & {.801} \\
    \rowcolor{gray!10} AirNet~\cite{li2022airnet} & 9M & 21.04 & {.884} & 32.98 & {.951} & 30.91 & {.882} & 24.35 & {.781} & 18.18 & {.735} & 25.49 & {.847} \\
    IDR~\cite{zhang2023ingredient} & 15M & {25.24} & {.943} & {35.63} & {.965} & \textcolor{tabred}{\textbf{31.60}} & \textcolor{tabred}{\textbf{.887}} & \textcolor{tabblue}{\textbf{27.87}} & \textcolor{tabblue}{\textbf{.846}} & 21.34 & {.826} & {28.34} & {.893} \\
    \rowcolor{gray!10} PromptIR~\cite{potlapalli2023promptir} & 36M & \textcolor{tabblue}{\textbf{26.54}} & \textcolor{tabblue}{\textbf{.949}} & \textcolor{tabblue}{\textbf{36.37}} & \textcolor{tabblue}{\textbf{.970}} & \textcolor{tabblue}{\textbf{31.47}} & \textcolor{tabblue}{\textbf{.886}} & \textcolor{tabred}{\textbf{28.71}} & \textcolor{tabred}{\textbf{.881}} & \textcolor{tabblue}{\textbf{22.68}} & \textcolor{tabblue}{\textbf{.832}} & \textcolor{tabblue}{\textbf{29.15}} & \textcolor{tabred}{\textbf{.904}} \\
    \midrule
     \ourmethod (\textit{ours}) & \textcolor{tabred}{\textbf{6M}} & \textcolor{tabred}{\textbf{29.84}} & \textcolor{tabred}{\textbf{.977}} & \textcolor{tabred}{\textbf{36.91}} & \textcolor{tabred}{\textbf{.977}} & 31.15 & {.882} & 26.86 & {.822} & \textcolor{tabred}{\textbf{23.50}} & \textcolor{tabred}{\textbf{.845}} & \textcolor{tabred}{\textbf{29.65}} & \textcolor{tabblue}{\textbf{.901}} \\
     \midrule 
     \multicolumn{14}{c}{Methods with the assistance of natural language prompts and large vision model} \\
     \midrule
     \rowcolor{gray!10}InstructIR~\cite{conde2024high} & 16M & 36.84 & .973 & 27.10 & .956 & 31.40 & .887 & 29.40 & .886 & 23.00 & .836 & 29.55 & .907 \\
     Perceive-IR~\cite{zhang2024perceive} & 42M & 28.19 & .964 & 37.25 & .977 & 31.44 & .887 &  29.46 & .886 & 22.88 & .833 & 29.84 &.909 \\
     \bottomrule
    \end{tabularx}
    \vspace{-3mm}
\end{table*}

\begin{table*}[t]
    \centering
    \footnotesize
    \fboxsep0.75pt  
    \setlength\tabcolsep{2.5pt}    
    \setlength{\extrarowheight}{0.2pt}
    \caption{\textit{Comparison to state-of-the-art for single degradations.} PSNR (dB, $\uparrow$) and SSIM ($\uparrow$) metrics are reported on the full RGB images. \textcolor{tabred}{\textbf{Best}} and \textcolor{tabblue}{\textbf{second best}} performances are highlighted. Our method excels prior work on dehazing and deraining. 
    }    
    \vspace{-3mm}
    \label{tab:exp:single}
    \begin{subtable}[l]{0.25\textwidth}
        \subcaption{\textit{Dehazing}}
        \begin{tabularx}{\textwidth}{X*{4}{c}}
        \toprule
        Method & Params. &\multicolumn{2}{c}{SOTS}\\
        \midrule
        \rowcolor{gray!10}DehazeNet~\cite{cai2016dehazenet}& - & 22.46 & .851\\
        MSCNN~\cite{ren2016single} & - &22.06&.908 \\
        \rowcolor{gray!10}AODNet~\cite{li2017aod} & - &20.29&.877\\
        EPDN ~\cite{qu2019enhanced} & - &22.57 & .863\\
        \rowcolor{gray!10}FDGAN~\cite{dong2020fdgan} & - &23.15 & .921 \\
        \midrule
        AirNet~\cite{li2022airnet} & 9M &23.18 & .900 \\
        \rowcolor{gray!10}PromptIR~\cite{potlapalli2023promptir}& 36M & \textcolor{tabred}{\textbf{31.31}} & \textcolor{tabblue}{\textbf{.973}} \\
        \midrule 
        \ourmethod (\textit{ours}) & \textcolor{tabred}{\textbf{6M}} & \textcolor{tabblue}{\textbf{30.92}} & \textcolor{tabred}{\textbf{.979}} \\
        \bottomrule
        \end{tabularx}
    \end{subtable}%
    \hfill
    \begin{subtable}[l]{0.25\textwidth}
        \subcaption{\textit{Deraining}}
        \begin{tabularx}{\textwidth}{X*{4}{c}}
        \toprule
        Method & Params. &\multicolumn{2}{c}{Rain100L}\\
        \midrule
        \rowcolor{gray!10}DIDMDN~\cite{zhang2018density}&-&23.79&.773\\
        UMR~\cite{yasarla2019uncertainty} & - & 32.39 & .921 \\
        \rowcolor{gray!10}SIRR~\cite{wei2019semi} & -& 32.37&.926 \\
        MSPFN~\cite{jiang2020multi} & - & 33.50 & .948 \\
        \rowcolor{gray!10}LPNet~\cite{gao2019dynamic}  & - &  23.15 & .921 \\
        \midrule
        AirNet~\cite{li2022airnet}  & 9M & 34.90 & .977 \\
        \rowcolor{gray!10}PromptIR~\cite{potlapalli2023promptir} & 36M & \textcolor{tabblue}{\textbf{37.04}} & \textcolor{tabblue}{\textbf{.979}} \\
        \midrule 
        \ourmethod (\textit{ours}) & \textcolor{tabred}{\textbf{6M}} & \textcolor{tabred}{\textbf{37.79}} & \textcolor{tabred}{\textbf{.982}}\\
        \bottomrule
        \end{tabularx}
    \end{subtable}%
    \hfill
    \begin{subtable}[l]{0.45\textwidth}
        \subcaption{\textit{Denoising}}
        \begin{tabularx}{\textwidth}{X*{7}{c}}
        \toprule
        Method & Params. & \multicolumn{2}{c}{$\sigma$=15} & \multicolumn{2}{c}{$\sigma$=25} & \multicolumn{2}{c}{$\sigma$=50}\\
        \midrule
        \rowcolor{gray!10}CBM3D~\cite{dabov2007color} & - & 33.50&.922&30.69&.868&27.36&.763 \\
        DnCNN~\cite{zhang2017beyond}&- & 33.89 & .930 & 31.23 & .883 & 27.92 & .789 \\
        \rowcolor{gray!10}IRCNN ~\cite{zhang2017learning}  & - & 33.87 & .929 & 31.18 & .882 & 27.88 & .790  \\
        FFDNet ~\cite{zhang2018ffdnet} & - & 33.87 & .929 & 31.21 & .882 & 27.96 & .789 \\
        \midrule 
        \rowcolor{gray!10}BRDNet~\cite{tian2000brdnet} & - & 34.10 & .929 & 31.43 & .885 & 28.16 & .794 \\ 
        AirNet~\cite{li2022airnet} & 9M & 34.14 & .936 & 31.48 & .893 & 28.23 & .806  \\
        \rowcolor{gray!10}PromptIR~\cite{potlapalli2023promptir} & 36M & \textcolor{tabred}{\textbf{34.34}} & \textcolor{tabred}{\textbf{.938}} & \textcolor{tabred}{\textbf{31.71}} & \textcolor{tabred}{\textbf{.897}} & \textcolor{tabred}{\textbf{28.49}} & \textcolor{tabred}{\textbf{.813}} \\
        \midrule 
        \ourmethod (\textit{ours})  & \textcolor{tabred}{\textbf{6M}} & \textcolor{tabblue}{\textbf{34.27}} & \textcolor{tabblue}{\textbf{.937}} & \textcolor{tabblue}{\textbf{31.63}} & \textcolor{tabblue}{\textbf{.896}} & \textcolor{tabblue}{\textbf{28.39}} &\textcolor{tabblue}{\textbf{.810}} \\
        \bottomrule
        \end{tabularx}
    \end{subtable}
    \vspace{-4mm}
\end{table*}

\noindent \textbf{Three Degradations.} We evaluate our All-in-One restorer, \ourmethod, against other specialized methods, including BRDNet~\cite{tian2020BRDnet}, LPNet~\cite{gao2019dynamic}, FDGAN~\cite{dong2020fdgan}, DL~\cite{fan2019dl}, MPRNet~\cite{zamir2021pmrnet}, AirNet~\cite{li2022airnet}, and PromptIR~\cite{potlapalli2023promptir}, all trained on three degradations: dehazing, deraining, and denoising. \ourmethod consistently leads as the top-performing All-in-One restorer, with an average gain of 0.45 dB across benchmarks and the most efficient architecture, as shown in \cref{tab:exp:3deg}. It achieves state-of-the-art results on SOTS and Rain100L, outperforming PromptIR by 0.38 dB and 1.83 dB, respectively. In denoising, \ourmethod ranks second across all $\sigma$ levels while maintaining 80\% fewer parameters and FLOPs.

\noindent \textbf{Five Degradations.} Extending the three degradation tasks to include deblurring and low-light enhancement~\cite{li2022airnet,zhang2023ingredient}, we validate our method’s comprehensive performance in an All-in-One setting. As shown in \cref{tab:exp:5deg}, \ourmethod effectively leverages degradation-specific features, surpassing AirNet~\cite{li2022airnet} and IDR~\cite{zhang2023ingredient} by an average of 4.16 dB and 1.22 dB, respectively, with 33\% and 60\% fewer parameters. Against general image restorers, \ourmethod outperforms Restormer~\cite{zamir2022restormer} and NAFNet~\cite{chen2022simple} on the LOLv1 dataset by 3.09 dB and 3.01 dB, while being four times smaller than Restormer and three times smaller than NAFNet.

\noindent \textbf{Single-Degradation.} We present results in \cref{tab:exp:single} with \ourmethod trained on individual degradation tasks: dehazing, deraining, and denoising. For dehazing, \ourmethod achieves the second-best PSNR and the highest SSIM. In deraining, \ourmethod surpasses the previous state-of-the-art, PromptIR~\cite{potlapalli2023promptir}, by 0.75 dB in PSNR and 0.003 in SSIM. For denoising, \ourmethod ranks second overall, effectively reducing noise while preserving details. 
Notably, while other models show significant performance drops in multi-degradation settings, \ourmethod maintains robust performance, attributed to its unified cross-degradation embedding. Furthermore, \ourmethod is substantially smaller than competing models, highlighting its efficiency and practical feasibility.

\noindent \textbf{Visual results.}
\begin{figure*}[t]
    \centering
    \includegraphics[width=0.97\linewidth]{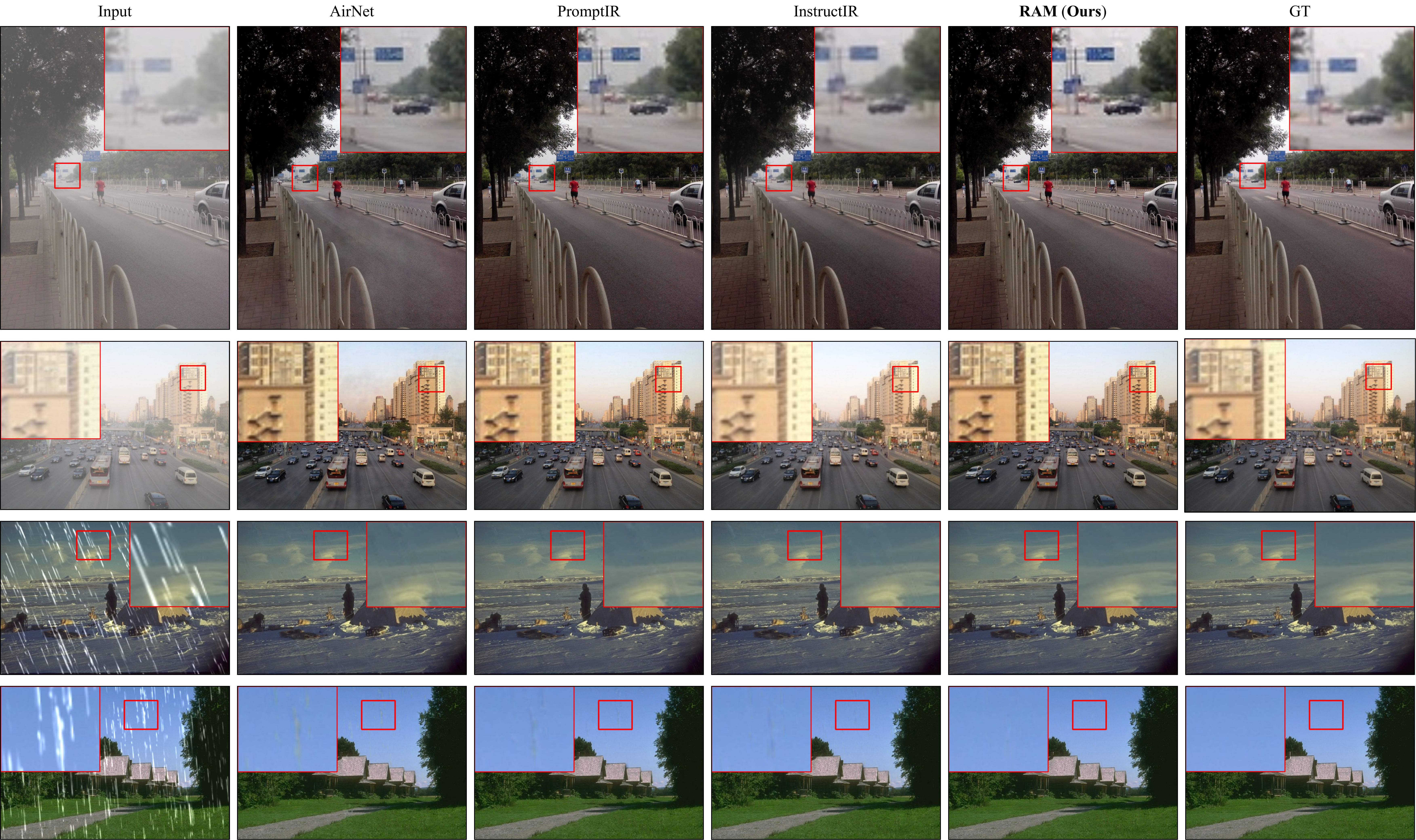}
    \vspace{-2mm}
    \caption{Visual comparison of \ourmethod with state-of-the-art methods considering three degradations. Zoom in for a better view.}
    \label{fig:exp:visual_results}
    \vspace{-1em}
\end{figure*}
To complement the quantitative results, we visualize the outcomes of our method in Fig.~\ref{fig:exp:visual_results}. The visualizations demonstrate the efficacy of \ourmethod across dehazing and draining, and we marked out the detailed region via the red boxes. The denoising visual results are provided in our \textit{Supp. Mat.} 
In the dehazing task, both AirNet~\cite{li2022airnet} and PromptIR~\cite{potlapalli2023promptir} exhibit limitations in fully eliminating haziness, leading to noticeable color reconstruction discrepancies. In contrast, our \ourmethod method effectively enhances visibility and ensures precise color reconstruction. 
In challenging rainy scenes, previous methods continue to exhibit remnants of rain streaks. Please, \textbf{zoom in} for more details. In contrast, our approach excels at eliminating these artifacts and recovering underlying details, showing our superiority under adverse weather conditions.
Note that the current SOTA PromptIR is approximately  6$\times$ larger than ours. Despite this, our method consistently produces visually superior results, demonstrating its effectiveness. For more detailed examples, please refer to the \textit{Supp. Mat.}

\noindent \textbf{Model efficiency.}
\begin{table}[t]
    \centering
    \footnotesize
    \setlength\tabcolsep{2.5pt}
    \setlength{\extrarowheight}{0.5pt}
    \caption{\textit{Complexity Analysis.} FLOPs are computed on an input image of size $224\times224$ using a NVIDIA Tesla A100 (40G) GPU.}
    \vspace{-3mm}
    \label{tab:efficiency_overview}
    \begin{tabularx}{\columnwidth}{X*{5}{c}}
        \toprule
        Method  &PSNR (dB, $\uparrow$)& Memory ($\downarrow$)  &Params. ($\downarrow$) & FLOPs ($\downarrow$) \\
        \midrule
        \rowcolor{gray!10}AirNet~\cite{li2022airnet} &31.20 &4829M &  8.93M  & 238G \\
        PromptIR~\cite{potlapalli2023promptir} & 32.06& 9830M & 35.59M & 132G \\
        \rowcolor{gray!10}IDR~\cite{zhang2023ingredient}&-& 4905M & 15.34M & 98G \\
        \midrule
        \ourmethod(\textit{ours}) &\textcolor{tabred}{\textbf{32.51}}&\textcolor{tabred}{\textbf{3556M}} &  \textcolor{tabred}{\textbf{6.29M}} & \textcolor{tabred}{\textbf{19G}} \\ 
        \bottomrule
    \end{tabularx}
    \vspace{-2.4em}
\end{table}

\cref{tab:efficiency_overview} compares memory usage, FLOPs, and model parameters, highlighting our framework’s efficiency over state-of-the-art all-in-one restorers. With a hybrid block design and our proposed degradation adaptation module, \ourmethod achieves a 0.45 dB PSNR improvement over PromptIR~\cite{potlapalli2023promptir} while using 63.8\% less memory, 82.3\% fewer parameters, and only 19G FLOPs, making it \textbf{85.6\%} more computationally efficient. This efficiency is due to our global-local intertwining in the sub-latent space, enabling both high performance and practical applicability. Our significant reductions in memory, parameters, and FLOPs position \ourmethod as a strong, efficient baseline for future research.

\begin{table}[t]
    \centering
    \footnotesize
    \caption{Ablation comparison of the effect of each component under the 3-degradation all-in-one task. The average PSNR is reported.}
    \vspace{-2mm}
    \label{tab:ablation}
    \setlength{\extrarowheight}{0.5pt}
    \setlength\tabcolsep{2pt}
    \scalebox{1}{
    \begin{tabularx}{\columnwidth}{X*{5}{c}}
        \toprule
        Method  &Skip-Split& Cross-Sigmod  &GatedDA & PSNR (dB, $\uparrow$) \\
        \midrule
        \rowcolor{gray!10} (a) & $\times$ & $\times$ & $\times$ & 30.85  \\
        (b) & \checkmark & $\times$ & $\times$ & 31.83  \\
        \rowcolor{gray!10} (c) & $\times$ & $\times$ & \checkmark & 32.13  \\
        (d) & \checkmark & $\times$ & \checkmark & 32.35  \\
        \rowcolor{gray!10} (e) & \checkmark & \checkmark & \checkmark & \textcolor{tabred}{\textbf{32.51}}  \\
        \bottomrule        
    \end{tabularx}
    }
    \vspace{-0.4em}
\end{table}

\begin{figure}[!t]
    \centering
    \includegraphics[width=1.0\linewidth]{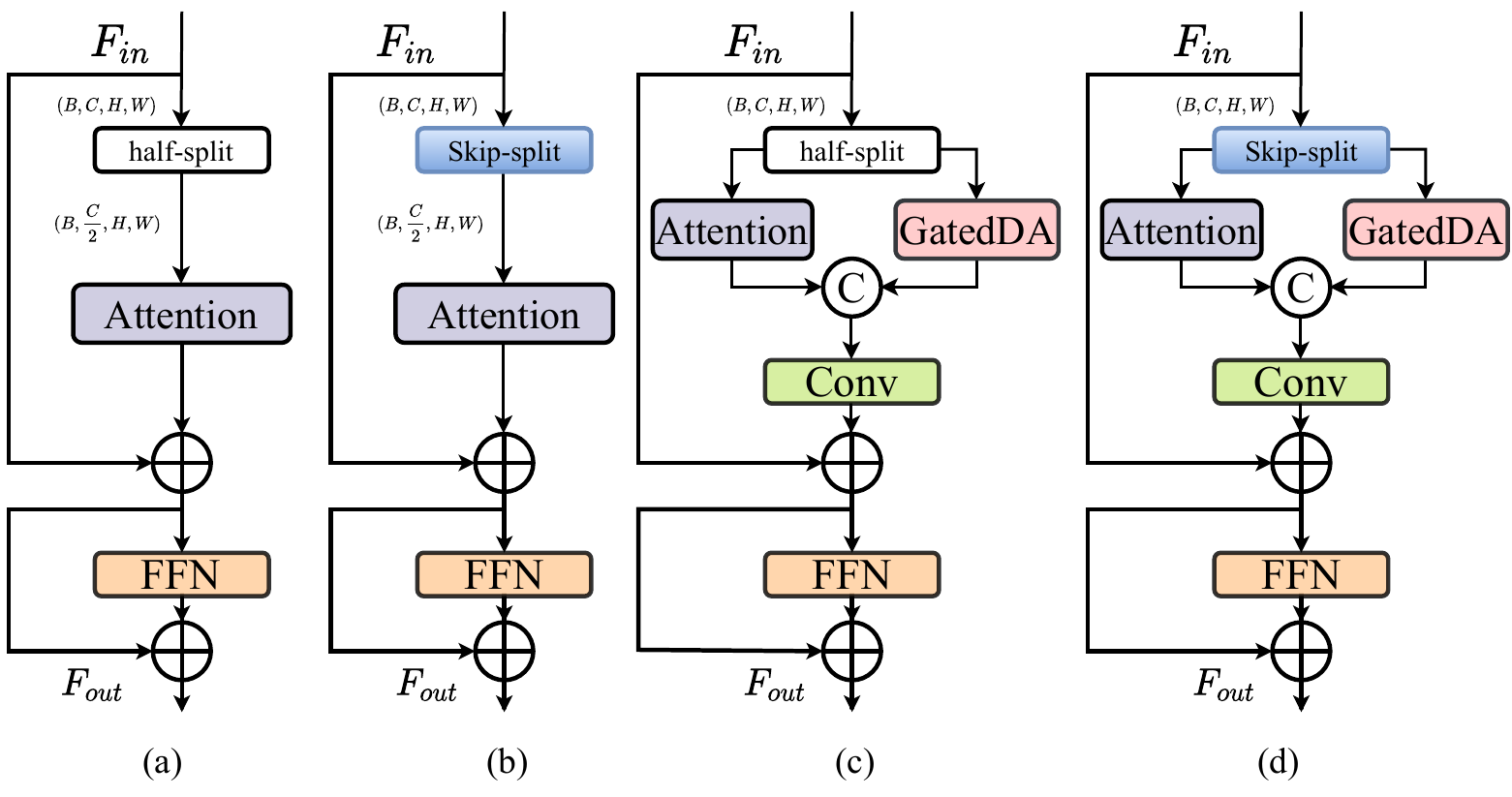}
    \vspace{-2em}
    \caption{Structure of different DAB variants.}
    \vspace{-1em}
    \label{fig:ab_component}
\end{figure}

\subsection{Ablation Studies}
\label{ablation}
\noindent \textbf{Impact of different components.} We conduct detailed studies on the proposed components within our \ourmethod framework. All experiments are conducted in the \textit{All-in-One} setting with \textit{three} degradations. We compare our plain feature modeling block DAB against other variants.  As detailed in \cref{tab:ablation}, we assess the effectiveness of our key architectural contributions by removing or replacing our designed module with the counterparts. The detailed architecture overview of these considered counterparts can be found in \cref{fig:ab_component}. 

We first examine the impact of our skip-split operation (\textit{a}-\textit{b}), which yields a significant improvement of 1.02 dB over the common half-split method. This validates and supports our motivation of deeply enabling the intertwining within the subparts of an input feature. Introducing our proposed GatedDA in parallel to the attention layer (\textit{c}) results in a substantial increase of $1.27$ dB. Combining GatedDA with the skip-split operation (\textit{d}) further enhances the reconstruction fidelity of our framework. This validates the effectiveness of our intention of using the local gated details to reconstruct the degradation-aware output. Lastly, the introduction of cross-feature filtering (\textit{e}), please refer to \cref{fig:framework} for our plain design, improves the interconnectivity between global-local features, further benefiting overall performance.

\begin{table}[t]
    \centering
    \footnotesize
    \setlength\tabcolsep{2.5pt}
    \setlength{\extrarowheight}{0.5pt}
    \caption{Results of different ($\alpha$, $\beta$, $\gamma$) for three-degradation.}
    \label{tab:ab_gatedDA}
    \vspace{-2mm}
    \begin{tabularx}{\columnwidth}{X*{3}{c}}
        \toprule
        ($\alpha$, $\beta$, $\gamma$)  &PSNR (dB, $\uparrow$)& SSIM ($\downarrow$)  \\
        \midrule
        \rowcolor{gray!10}(0, \nicefrac{hidden}{2}, \nicefrac{hidden}{2})  & 32.14 & {.910} \\
        (\nicefrac{hidden}{2}, 0, \nicefrac{hidden}{2}) &32.21 &{.912}  \\
        \rowcolor{gray!10}(\nicefrac{hidden}{2}, \nicefrac{hidden}{2}, 0) &31.37 &{.907}  \\
        (\nicefrac{hidden}{4}, \nicefrac{hidden}{4}, \nicefrac{hidden}{2}) &\textcolor{tabred}{\textbf{32.51}} & \textbf{\textcolor{tabred}{.916}}  \\ 
        \bottomrule
    \end{tabularx}
    \vspace{-2em}
\end{table}

\begin{figure}[!t]
    \centering
    \includegraphics[width=1.0\linewidth]{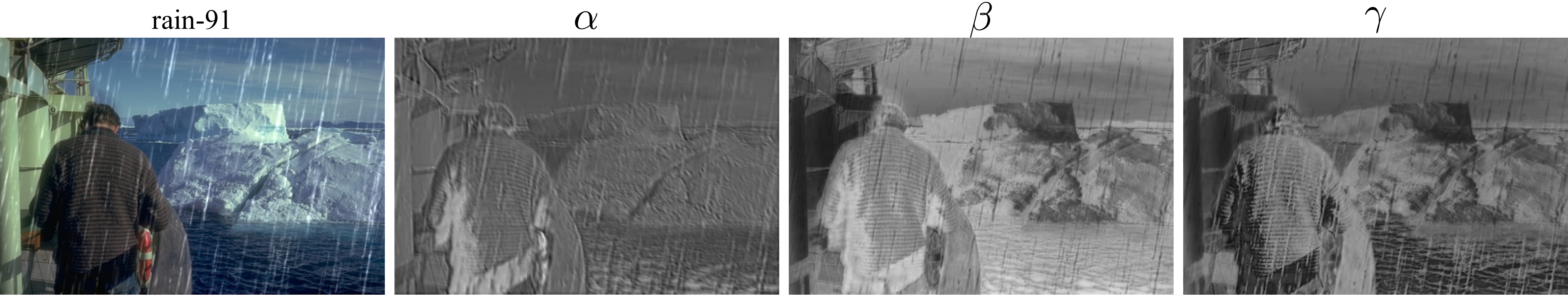}
    \vspace{-2em}
    \caption{Visual feature maps of $\alpha$, $\beta$, and $\gamma$ within GatedDA.}
    \vspace{-1em}
    \label{fig:ab_gic}
\end{figure}

\noindent \textbf{Impact of different $\alpha$, $\beta$, and $\gamma$ values in Gated DA.} Tab.\ref{tab:ab_gatedDA} shows the different impact of different value of ($\alpha$, $\beta$, $\gamma$) in the proposed gatedDA. We noticed that when setting one of these parameters to 0, the performance is decreased. When we set ($\alpha$, $\beta$, $\gamma$) to ($\nicefrac{hidden}{4}$, $\nicefrac{hidden}{4}$, $\nicefrac{hidden}{2}$), the average performance is increased. This means each part of the proposed gatedDA is necessary. The visual features shown in \cref{fig:ab_gic} indicate that $\alpha$, $\beta$, and $\gamma$ consistently focus on various aspects of the degraded regions, each specializing in different levels or types of degradation.

\begin{figure}[!t]
    \centering
    \includegraphics[width=1.0\linewidth]{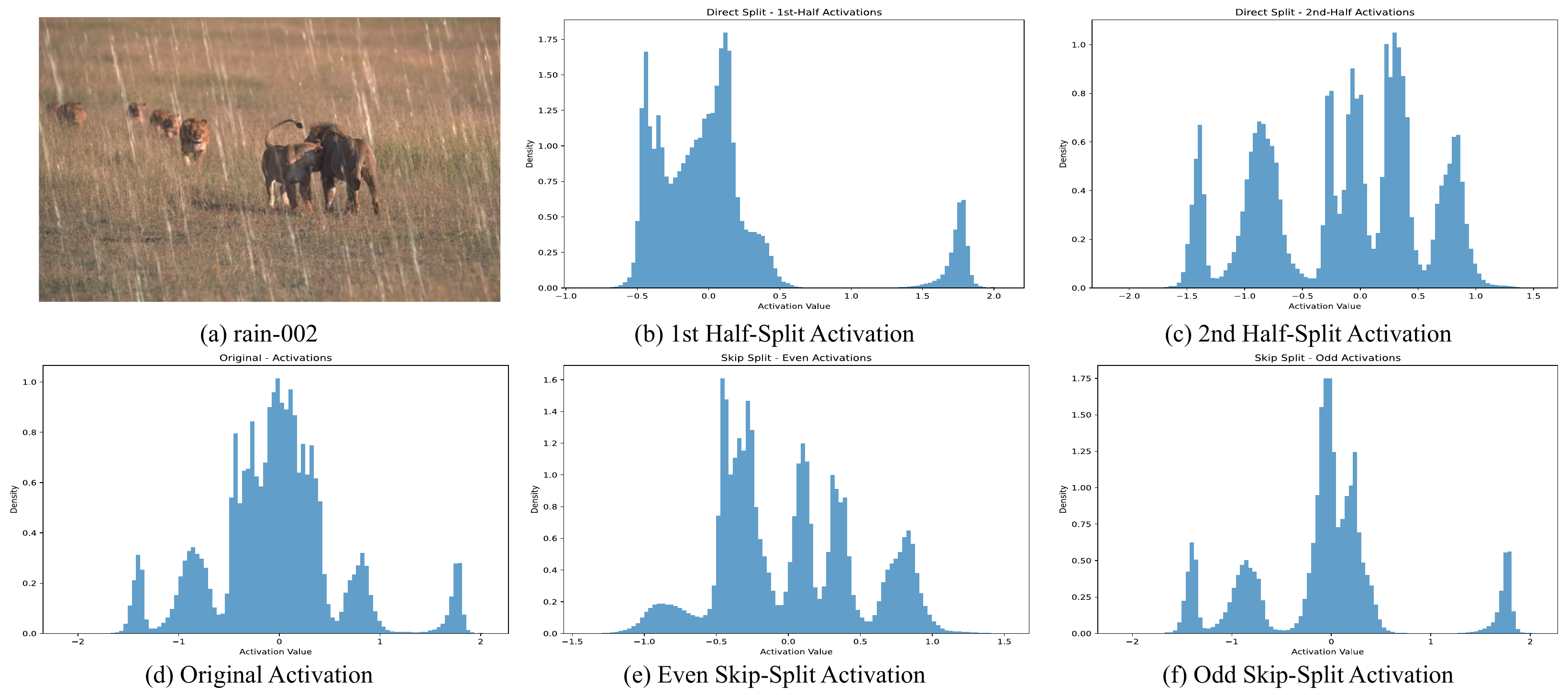}
    \vspace{-2em}
    \caption{Visulization of the channel activation distribution.}
    \vspace{-1em}
    \label{fig:vis_skip}
\end{figure}
\section{Discussion}
\noindent\textbf{What does the proposed Skip-Split bring?}
Adjacent channels often contain redundant information due to spatial correlations in the data. Directly splitting the channels into two contiguous halves can lead to uneven feature distribution, with one half potentially capturing redundant features while the other lacks important information. By using a simple skip-split method that interleaves channels between the two processing paths, we ensure a more balanced and diverse set of features in each path, enhancing the effectiveness of both the self-attention and convolutional components. These phenomena is also validated in \cref{fig:vis_skip}.

\begin{figure}[!t]
    \centering
    \includegraphics[width=1.0\linewidth]{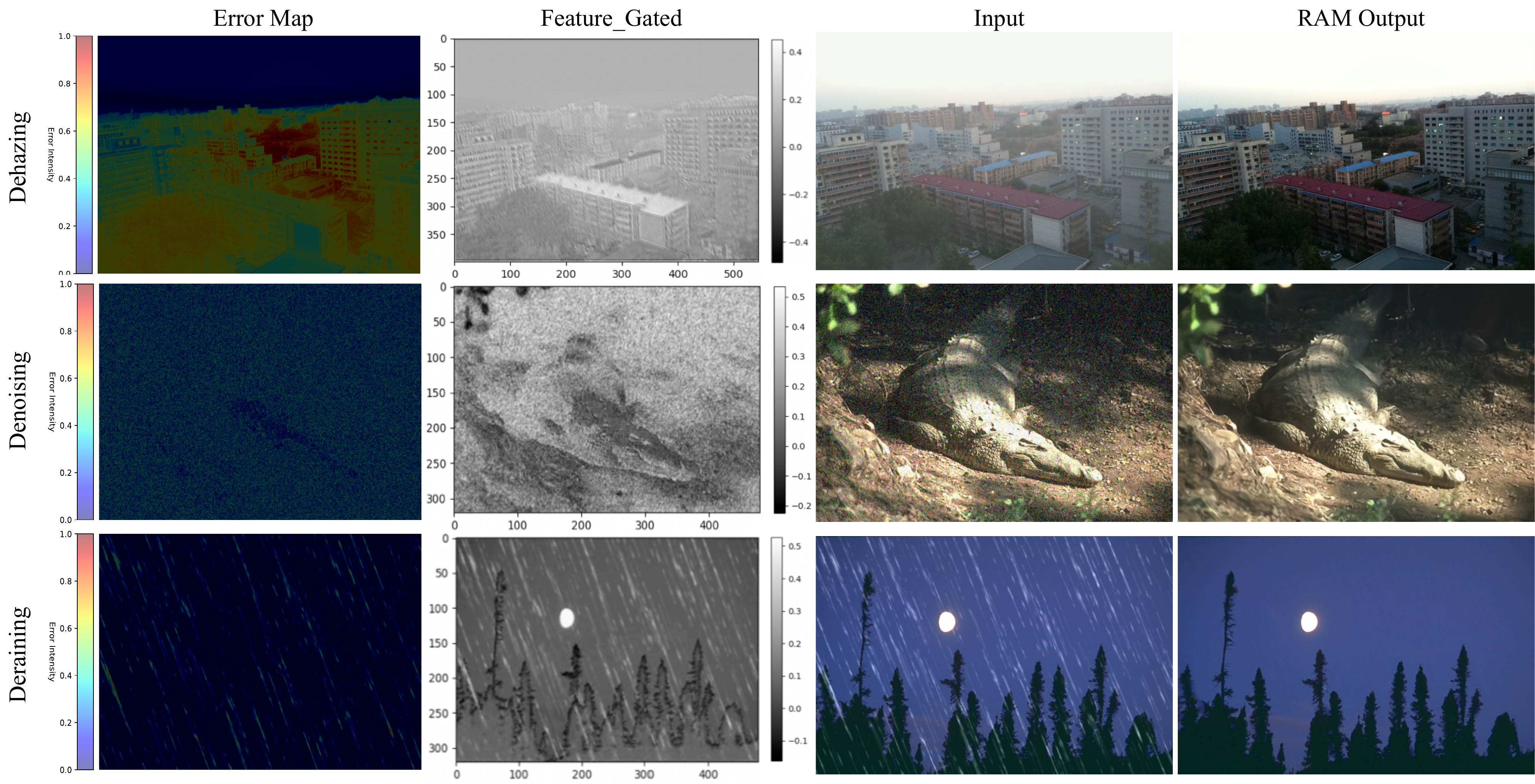}
    \vspace{-2em}
    \caption{Visulization of the error map and the output of GatedDA.}
    \vspace{-1.5em}
    \label{fig:vis_gated}
\end{figure}

\noindent\textbf{Why does the proposed GatedDA work?} 
As shown in \cref{fig:vis_gated}, the error map visualization reveals that degradation in an input image is often not evenly distributed, instead presenting as both global patterns and localized clusters. This suggests that effective degradation modeling must capture both widespread and specific distortions. The proposed GatedDA module addresses this need by targeting its feature responses specifically on degraded regions, achieving a close alignment with the error map. This focus underscores GatedDA’s capacity to selectively enhance degradation features, and when combined with global attention, it provides a comprehensive representation of the image's degradation structure. This synergy between GatedDA and global attention improves overall restoration quality by allowing nuanced handling of various degradation types. Importantly, our results demonstrate that GatedDA maintains this adaptability across diverse degradation conditions, emphasizing its robustness and versatility in restoration tasks.

\noindent\textbf{Why is \ourmethod efficient?} \ourmethod reduces computational costs by splitting input channels: one half is processed by self-attention, the other by a Gated CNN block. This division halves the self-attention’s complexity from \(O(B \cdot \text{head} \cdot (H \cdot W)^2)\), while the Gated CNN processes the remaining channels with a lower complexity of \(O(B \cdot C \cdot H \cdot W)\), especially beneficial for high-resolution inputs where \((H \cdot W)^2\) dominates. Additionally, \ourmethod employs dimensionality reduction and parameter-efficient designs, collectively reducing GFLOPs and model parameters. This balance of global context modeling and efficient local feature extraction enables \ourmethod to minimize computational costs without sacrificing performance.

\noindent\textbf{Is scaling down the new advantage?}  
While recent methods in image restoration often scale up model size and complexity, our approach takes a different path: scaling down. Instead of relying on large architectures, we embrace a simple but non-trivial design, using targeted components like the GatedDA module to effectively capture degradation without excessive parameters. This simplicity is not a compromise; it’s an asset—yielding high-quality restoration with minimal computational demand, faster training, and greater adaptability.
The All-in-One IR framework, though new compared to degradation-specific methods, faces a key limitation: an imbalance in data distribution, with certain degradations (\eg, dehazing) dominating. Our experiments suggest that a more balanced dataset could significantly improve performance across degradation types, offering a constructive direction for future work. Additional analyses, experiments, and visual results are available in our \textit{Supp. Mat.}
\section{Conclusion}
\label{sec:conclusion}
We introduce \ourmethod, an efficient single-image restoration model capable of handling any type of degradation. Our model leverages the intrinsic characteristics of each degradation as gated guidance, directing the overall feature modeling process. This approach enables the joint embedding of all degradation types while preserving each of their unique attributes. \ourmethod sets a new standard in degradation awareness, achieving unparalleled efficiency and fidelity in all-in-one image restoration. We hope that our model will serve as a fresh baseline for future research in this field.

{
    \small
    \bibliographystyle{ieeenat_fullname}
    \bibliography{reference}
}

\end{document}